\documentclass[conference]{IEEEtran}
\IEEEoverridecommandlockouts
\usepackage{cite}
\usepackage{amsmath,amssymb,amsfonts}
\usepackage{algorithmic}
\usepackage{graphicx}
\usepackage{pgfplots}
\pgfplotsset{compat=1.14}
\usepackage{textcomp}
\usepackage{multirow}

\begin{document}

\title{Neuromorphic hardware as a self-organizing computing system}

\author{
\IEEEauthorblockN{Lyes Khacef$^1$, Bernard Girau$^2$, Nicolas Rougier$^3$, Andres Upegui$^4$, Beno\^it Miramond$^{1,*}$}
\IEEEauthorblockA{\textit{$^1$ University C\^ote d'Azur - LEAT / CNRS UMR 7248}
\textit{$^2$ University of Lorraine - Loria} \\
\textit{$^3$ Inria Bordeaux Sud-Ouest - LaBRI / University of Bordeaux / CNRS UMR 5800} \\
\textit{$^4$ University of Applied Sciences of Western Switzerland  InIT - hepia - HES-SO} \\
$^*$ Corresponding author: benoit.miramond@unice.fr}}

\maketitle

\begin{abstract}
This paper presents the self-organized neuromorphic architecture named SOMA. The objective is to study neural-based self-organization in computing systems and to prove the feasibility of a self-organizing hardware structure. Considering that these properties emerge from large scale and fully connected neural maps, we will focus on the definition of a self-organizing hardware architecture based on digital spiking neurons that offer hardware efficiency. From a biological point of view, this corresponds to a combination of the so-called synaptic and structural plasticities. We intend to define computational models able to simultaneously self-organize at both computation and communication levels, and we want these models to be hardware-compliant, fault tolerant and scalable by means of a neuro-cellular structure.
\end{abstract}

\section{Introduction}

Several current issues such as analysis and classification of major data sources (sensor fusion, Internet of Things, etc.), and the need for adaptability in many application areas (automotive systems, autonomous drones, space exploration, etc.), lead us to study a desirable property from the brain that encompasses all others: the cortical plasticity.
This term refers to one of the main developmental properties of the brain where the organization of its structure (structural plasticity) and the learning of the environment (synaptic plasticity) develop simultaneously toward an optimal computing efficiency.
Such developmental process is only made possible by some key features: focus on relevant information, representation of information in a sparse manner, distributed data processing and organization fitting the nature of data, leading to a better efficiency and robustness.
Our goal is to understand and design the first artificial blocks that are involved in these principles of plasticity.
Hence, transposing plasticity, and its underlying blocks, into hardware will contribute to define a substrate of computation endowed with self-organization properties stemming from the learning of incoming data.

Neural principles of plasticity may not be sufficient to ensure that such a substrate of computation is scalable enough in the perspective of future massively parallel an distributed devices. Our claim is that the expected properties of such alternative computing devices could emerge from a close interaction between cellular computing (decentralization and hardware compliant massive parallelism) and neural computation (self-organization and adaptation). We also claim that neuro-cellular algorithmics and hardware design are so tightly related that these two aspects should be studied together.
Therefore we propose to combine neural adaptativity and cellular computing efficiency through a neuro-cellular approach of synaptic and structural self-organization that defines a fully decentralized control layer for neuromorphic reconfigurable hardware.

For this aim, the project gathers neuroscientists, computer science researchers, hardware architects and micro-electronics designers to explore the concepts of a Self-Organizing Machine Architecture: SOMA.
This Self-Organization property already studied in various fields of computer science (artificial neural networks, multi-agents and swarm systems, cellular automata, etc.), is studied for the very first time in a new context with a transverse look from the computational neuroscience discipline to the design of reconfigurable microelectronic circuits.
The project focuses on the blocks that will pave the way in the long term for smart computing substrates, exceeding the limits of current technology.
The SOMA architecture will practically define an original brain-inspired computing system that will be prototyped onto FPGA devices.


\section{Self-organizing neural models}
\label{sec:models}
\begin{figure}[!h]
\begin{minipage}{0.5\linewidth}
  \centerline{\includegraphics[width=0.9\linewidth]{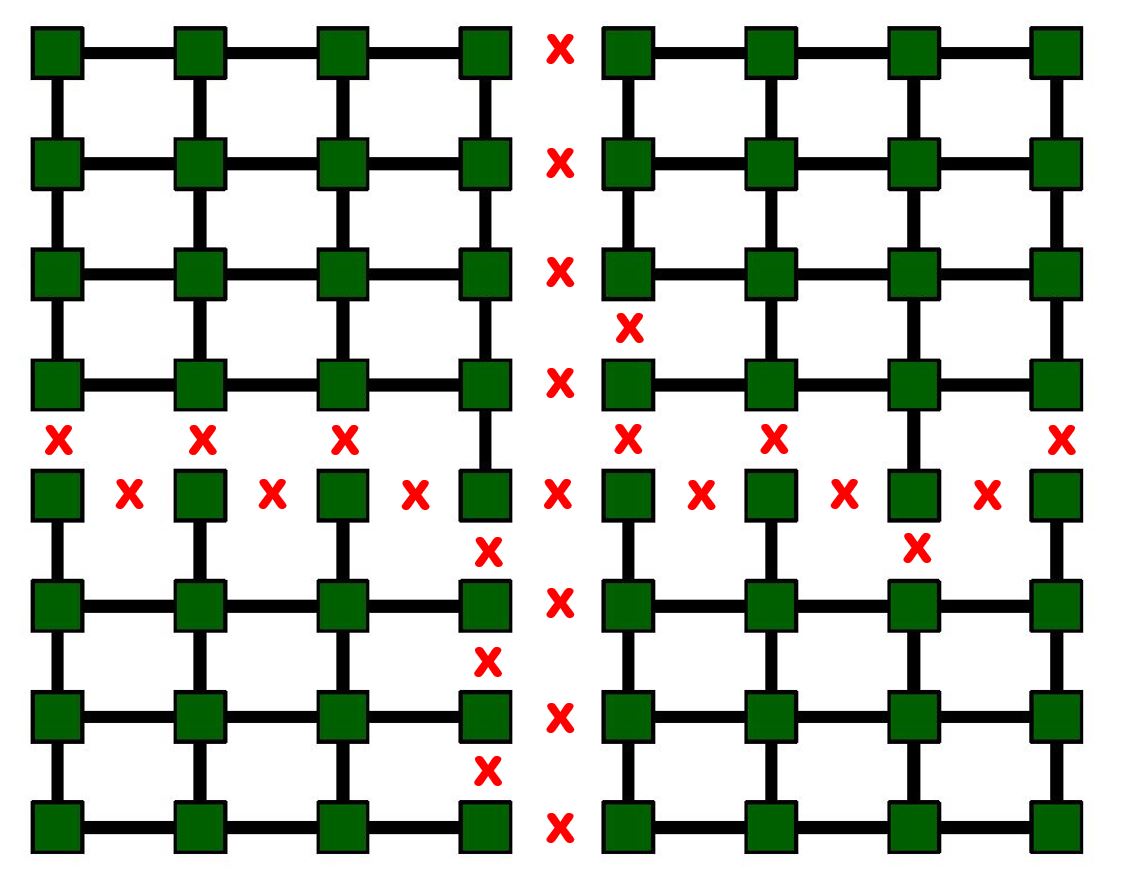}}
\end{minipage}%
\begin{minipage}{0.5\linewidth}
  \centerline{\includegraphics[width=0.9\linewidth]{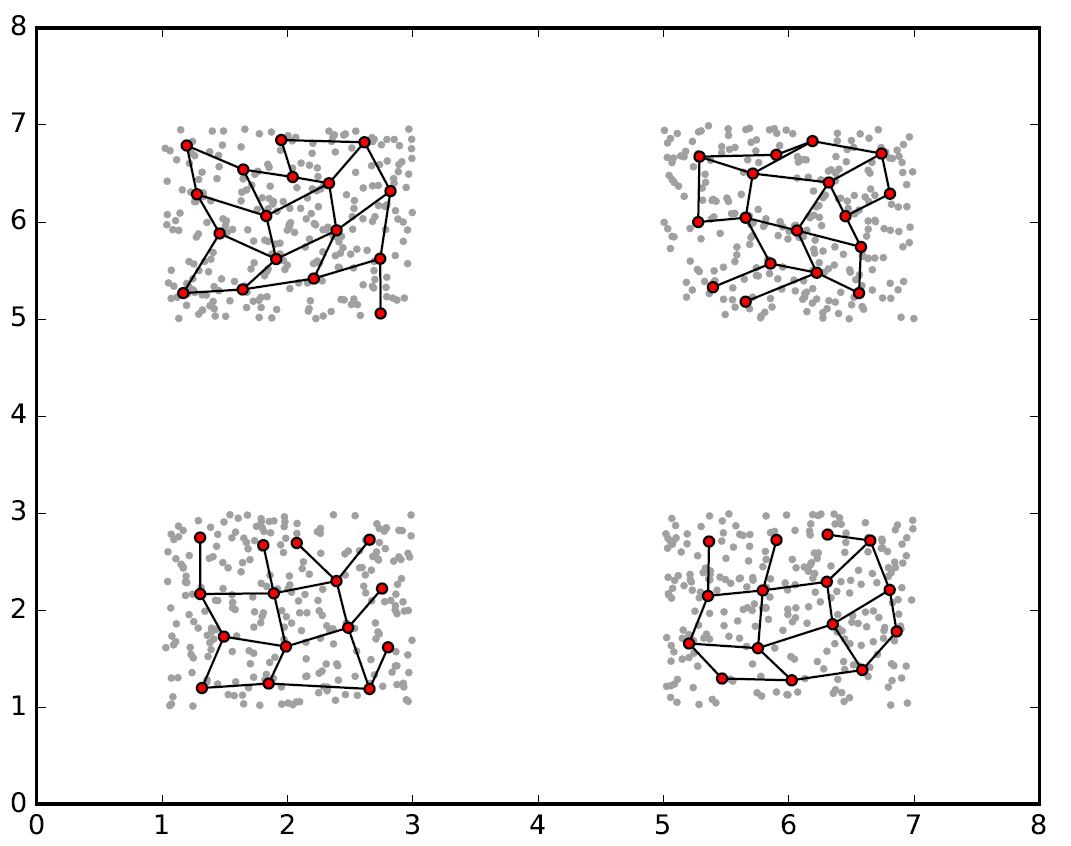}}
\end{minipage}
\caption{(Left) Cellular architecture connections after training with $w = 3e - 07$. (Right) Weights and probability density after training with $w = 3e - 07$.}
\label{psom}
\end{figure}

The gap between existing fixed computing systems and dynamic self-organized substrates may be filled with the help of computational neuroscience through the definition of neural models that exhibit properties like unsupervised learning, self-adaptation, self-organization, and fault tolerance which are of particular interest for efficient computing in embedded and autonomous systems.
However, these properties only emerge from large fully connected neural maps that result in intensive synaptic communications.
Previous works have for example already showed the possibility of using neural self-organizing models to control task allocation in manycore substrates \cite{rod013}.
Other works have also proposed adaptation of neural computational paradigms to cellular (neighborhood) constraints (DMAD-SOM \cite{rodriguez2015}, RSDNF\cite{chappetdevangel:hal-01071862}, CASAS\cite{chappetdevangel:hal-01264903}). 
Previous works have also studied different approaches to define cellular self-reconfiguration \cite{DBLP:journals/biosystems/StaufferMRV08} and self-organization\cite{DBLP:journals/ijrc/GirauTVB09} onto FPGA-based hardware.
This paper addresses the challenge of defining a neural model supporting hardware self-organization by lying at the intersection of four main research fields, namely adaptive reconfigurable computing, cellular computing, computational neuroscience, and hardware neurocomputing.
We thus propose a dynamically laterally connected neural model (map) that permits us to think about modifying the network topology in order to better fit the probability density function of incoming data. Synaptic pruning and sprouting are two biological mechanisms permitting structural plasticity. In the case of the model presented here, we are only using pruning by defining a probability for a lateral synaptic connection to be removed. A useless synapse connects two neurons whose activities are poorly correlated. Such poor correlation is expressed as a high distance between the weight vectors representing both neurons, and a very low activity (winning) of neurons which reflects a poor capability of the prototype vector to represent the probability density function. All these factors are modulated by a pruning rate $w$ and determine the probability of a given synapse to be removed as illustrated in Figure~\ref{psom}.


\section{Self-organizing machine architecture}
The neural-based mechanism responsible of the self-organization is integrated into a cellular processing architecture. It is decomposed in four distinct layers intended to provide the hardware plasticity targeted by the SOMA project. These layers are: (1) data acquisition, (2) pre-processing, which can be in the form of feature extraction, (3) self-organization of computation and communications, and (4) computation in the form of a reconfigurable computation unit (FPGA or processor). These four  architecture layers have been presented in \cite{rodriguez2015} and we already designed preliminary versions of the three first layers in \cite{rod013,fiackAHS2014}.

The first results obtained with this architecture showed that the system was able to solve a task allocation problem in a distributed way. Yet, the means to achieve it are not completely satisfactory for several reasons. First, the system architecture, although adaptive, relies on a neural model whose structure is fixed a priori and often results in expensive communication times.
Secondly, the architecture does not implement a distributed clustering algorithm, which makes it difficult to classify a node into a computing area once the neural network has learned the data from the environment.

Finally, the adaptation is based on the temporal correlations between the modalities in the input space, which leads to a considerable increase in the number of clusters (areas) in the computation architecture.
The aim of this paper is to avoid the issues mentioned above by proposing an evolution of the existing adaptive methods. This method takes advantage of synaptic pruning and will lighten the cost of communication over the learning process to isolate the neurons belonging to the same cluster. The third problem could be raised by hierarchically constructing a single neural map per data type and a upper map for the merge of the different modalities.

\begin{figure}[!h]
	\centerline{\includegraphics[width=0.9\linewidth]{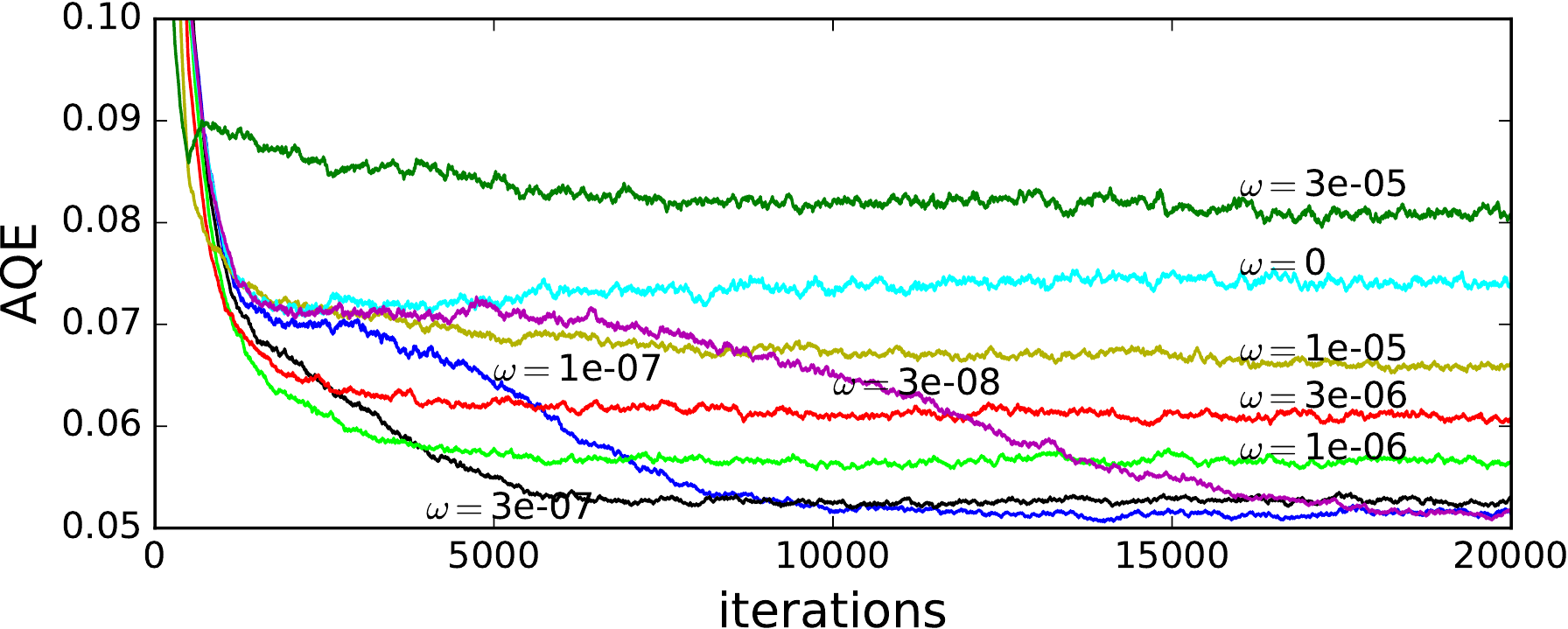}}
	\caption{AQE vs. time for different pruning rates $w$. $w \neq 0$ corresponds to a dynamic neural map with pruning enabled.}
	\label{aqe}
\end{figure}

The proposed self-organizing mechanisms will be exploited by user-defined applications running on a multicore array in the form of a NoC-based manycore system. 
The NoC architecture we currently use is based on the HERMES routing architecture~\cite{moraes2004hermes} for inter-node communication, for which we previously proposed an adapted version of the NoC architecture to support dynamic reconfiguration \cite{fiackAHS2014}. The main novelty concerns the coupling with the self-organizing mechanism.
The routing layer implements the neural model briefly presented in section \ref{sec:models}.
The model has been adapted for cellular architectures and integrates pruning capabilities. We applied the model to a quantization problem where we try to adapt the organization of the hardware to a density function representing the environment as illustrated in Figure~\ref{psom}. Preliminary results have shown that the method removes useless lateral connections in order to better fit the target probability density functions. The results showed that the proposed pruning mechanisms may improve the network performance by reducing the average quantization error of the incoming stimuli as illustrated in Figure~\ref{aqe}.
In our future works, the self-organizing layer will be responsible for the coordination of nodes in order to decide which node must deal with new incoming data in a completely decentralized manner. It will also be able to reconfigure the computing layer in order to better fit the hardware to the nature of incoming data. 

\section{Conclusion}
Hence, this paper proposed a convergence point between past research approaches toward new computation paradigms based on adaptive reconfigurable architecture and neuromorphic hardware.
The presented SOMA architecture is based on cellular computing and targets a massively parallel, distributed and decentralized neuromorphic architecture.
The self-organizing mechanism integrated into the architecture is based on neural models inspired from brain plasticity where the hardware organization emerges from the interactions between neural maps transposed into hardware. The first results presented in this paper have shown how such dynamical structure can adapt to random data representing a dynamic and evolving environment.

\bibliographystyle{IEEEtran}
\footnotesize
\bibliography{biblio}
\end{document}